\documentclass[11pt,a4paper]{article}
\usepackage[hyperref]{eacl2021}
\usepackage{times}
\usepackage{latexsym}

\usepackage{graphicx}

\usepackage{multirow}

\usepackage{amssymb}

\usepackage{float}
\restylefloat{table}

\usepackage{verbatim}

\usepackage{tabularx}

\usepackage{microtype}

\aclfinalcopy 

\title{MultiWOZ 2.3: A multi-domain task-oriented dialogue dataset enhanced with annotation corrections and co-reference annotation}

\author{Ting Han$^{1}$\footnotemark[1] \quad Ximing Liu$^2$\footnotemark[1] \quad Ryuichi Takanobu$^3$ \quad Yixin Lian$^2$ \\
\textbf{Chongxuan Huang$^2$ \quad Dazhen Wan$^3$ \quad Wei Peng$^{2}\footnotemark[2]$ \quad Minlie Huang$^{3}\footnotemark[2]$ }\\
  $^1$University of Illinois at Chicago\\
  $^2$Artificial Intelligence Application Research Center (AARC), Huawei Technologies\\
  $^3$Tsinghua University \\
  \texttt{than24@uic.edu, aihuang@tsinghua.edu.cn}\\
  \texttt{\{gxly19, wandz19\}@mails.tsinghua.edu.cn}\\
  \texttt{\{liuximing1, huang.chongxuan, lianyixin1, peng.wei1\}@huawei.com}\\}

\date{}

\begin{document}
\maketitle

\renewcommand{\thefootnote}{\fnsymbol{footnote}}
\footnotetext[1]{Both authors contributed equally to the work. The work was conducted when Ting Han was interning at Huawei AARC.}
\footnotetext[2]{Corresponding author.}

\begin{abstract}

Task-oriented dialogue systems have made unprecedented progress with multiple state-of-the-art (SOTA) models underpinned by a number of publicly available MultiWOZ datasets. Dialogue state annotations are error-prone, leading to sub-optimal performance. Various efforts have been put in rectifying the annotation errors presented in the original MultiWOZ dataset. In this paper, we introduce MultiWOZ 2.3, in which we differentiate incorrect annotations in dialogue acts from dialogue states, identifying a lack of co-reference when publishing the updated dataset. To ensure consistency between dialogue acts and dialogue states, we implement co-reference features and unify annotations of dialogue acts and dialogue states. We update the state of the art performance  of natural language understanding and dialogue state tracking on MultiWOZ 2.3, where the results show significant improvements than on previous versions of MultiWOZ datasets (2.0-2.2).

\end{abstract}

\section{Introduction}

\renewcommand{\thefootnote}{\arabic{footnote}}

Task-oriented dialogue systems have made unprecedented progress with multiple state-of-the-art (SOTA) models underpinned by a number of publicly available datasets ~\citep{zhu2020crosswoz, henderson_etal_2014_second, williams_etal_2013_dialog, wen_etal_2017_network, rastogi2019towards, budzianowski2018large}.


As the first publicly released dataset, MultiWOZ hosts more than 10K dialogues across eight different domains covering ``Train", ``Taxi",  ``Hotel",  ``Restaurant",  ``Attraction",  ``Hospital",  ``Bus" and ``Police".  MultiWOZ has been widely adopted by researchers in dialogue policy~\citep{takanobu_etal_2019_guided, zhao_etal_2019_rethinking}, dialogue generation~\citep{chen_etal_2019_semantically} and dialogue state tracking ~\citep{zhou2019multi, zhang2019find, heck_etal_2020_trippy, lee_etal_2019_sumbt, wu_etal_2019_transferable} as it provides a means for modeling the changing states of dialogue goals in multi-domain interactions.

Dialogue state annotations are error-prone, leading to sub-optimal performance. For example, the SOTA joint accuracy for dialogue state tracking (DST) is still below or around  60\%.\footnote[1]{\href{https://github.com/budzianowski/multiwoz}{https://github.com/budzianowski/multiwoz}.Marked date: 6-1-2021} MultiWOZ 2.1~\citep{eric_etal_2020_multiwoz} was released to rectify annotation errors presented in the original MultiWOZ dataset. MultiWOZ 2.1 introduced additional features such as slot descriptions and dialogue act annotations for both systems and users via ConvLab~\citep{lee_etal_2019_convlab}. Further efforts have been put into MultiWOZ 2.2~\citep{zang_etal_2020_multiwoz} to improve annotation quality. This schema-based dataset contains annotations allowing for directly retrieving slot values from a given dialogue context ~\citep{zhang2019find, gao_etal_2019_dialog, heck_etal_2020_trippy}. Despite achieving a noticeable annotation quality uplift compared to that for the original MultiWOZ, there is still room to improve. The focus of the corrections is on dialogue state annotations leaving the problematic dialogue act annotations untouched. Furthermore, the critical co-reference and ellipsis feature prevalent in the human utterance is not in presence.

\begin{table*}[!htb]
\resizebox{\textwidth}{50mm}{
\begin{tabular}{|l|l|l|l|l|}
\hline
\textbf{Error Type} & \textbf{Dialogue ID} & \textbf{Utterance} & \textbf{2.1 Dialog\_act} & \textbf{2.3 Dialog\_act}  \\ \hline

\multirow{3}{10mm}{Under-annotated}   & SSNG0348.json & \begin{tabular}[c]{@{}l@{}}For 3 people starting on Wednesday\\ and staying 2 nights .\end{tabular}                                                               & Hotel-Inform.Stay: 2                                                                                                                                                                           & \begin{tabular}[c]{@{}l@{}}Hotel-Inform.Stay: 2\\ {\color{red}Hotel-Inform.Day: Wednesday}\\ {\color{red}Hotel-Inform.Day: 3}\end{tabular}                                                     \\ \cline{2-5}
 
                                                                              & PMUL1170.json & \begin{tabular}[c]{@{}l@{}}Yes , one ticket please , can I also get\\ the reference number ?\end{tabular}                                                         & Train-Inform.People: {\color{red}1}                                                                                                                                                                         & \begin{tabular}[c]{@{}l@{}}Train-Inform.People: {\color{red}one}\\ {\color{red}Train-Request.Ref: ?}\end{tabular}                                                                              \\ \cline{2-5}

                                                                              & SNG01856.json & \begin{tabular}[c]{@{}l@{}}no, i just need to make sure it's cheap.\\ oh, and i need parking\end{tabular}                                                         & Hotel-Inform.Parking: yes                                                                                                                                                                      & \begin{tabular}[c]{@{}l@{}}Hotel-Inform.Parking: yes\\ {\color{red}Hotel-Inform.Price: cheap}\end{tabular}                                                                        \\ \hline

\multirow{3}{10mm}{Wrongly-annotated} & PMUL2596.json & \begin{tabular}[c]{@{}l@{}}I will need to be picked up at the hotel\\ by 4:45 to arrive at the college on\\ tuesday .\end{tabular}                                & \begin{tabular}[c]{@{}l@{}}Taxi-Inform.Leave:{\color{red} 04:45}\\ Taxi-Inform.Depart: {\color{red}arbury}\\ {\color{red}lodge guesthouse}\\ Hotel-Inform.Day: tuesday\end{tabular}                                                   & \begin{tabular}[c]{@{}l@{}}Taxi-Inform.Leave: {\color{red}4:45}\\ {\color{red}Taxi-Inform.Dest: the college}\\ Taxi-Inform.Depart: {\color{red}the hotel}\\ Hotel-Inform.Day: tuesday\end{tabular}          \\ \cline{2-5}

                                                                              & PMUL3296.json & \begin{tabular}[c]{@{}l@{}}Yeah , could you book me a room for 2\\ people for 4 nights starting Tuesday ?\end{tabular}                                            & \begin{tabular}[c]{@{}l@{}}Hotel-Inform.Stay: {\color{red}2}\\ Hotel-Inform.Day: Tuesday\\ Hotel-Inform.People:{\color{red} 4}\end{tabular}                                                                              & \begin{tabular}[c]{@{}l@{}}Hotel-Inform.Stay: {\color{red}4}\\ Hotel-Inform.Day: Tuesday\\ Hotel-Inform.People:{\color{red} 2}\end{tabular}                                                    \\ \cline{2-5}

                                                                              & PMUL4899.json & \begin{tabular}[c]{@{}l@{}}How about funkyu fun house , the are\\ located at 8 mercers row , mercers ro\\ industrial estate .\end{tabular}                        & \begin{tabular}[c]{@{}l@{}}Attraction-Recommend.Name:\\ funky fun house\\ {\color{red}Attraction-Recommend.Addr: 8}\\ {\color{red}mercers row}\\ {\color{red}Attraction-Recommend.Addr:}\\ {\color{red}mercers row industrial estate}\end{tabular} & \begin{tabular}[c]{@{}l@{}}Attraction-Recommend.Name:\\ funky fun house\\ {\color{red}Attraction-Recommend.Addr: 8}\\ {\color{red}mercers row , mercers row} \\ {\color{red}industrial estate}\end{tabular} \\ \hline

\multirow{3}{10mm}{Over-annotated}    & PMUL3250.json & \begin{tabular}[c]{@{}l@{}}No , I apoligize there are no Australian\\ restaurants in Cambridege . Would you \\ like to try another type of cuisine ?\end{tabular} & \begin{tabular}[c]{@{}l@{}}Restaurant-Request.Food: ?\\ Restaurant-NoOffer.Food: \\ Australina\\ {\color{red}Restaurant-NoOffer.Area:}\\ {\color{red}Cambridge}\end{tabular}                                             & \begin{tabular}[c]{@{}l@{}}Restaurant-Request.Food: ?\\ Restaurant-NoOffer.Food:\\ Australian\end{tabular}                                                           \\ \cline{2-5}

                                                                              & MUL1118.json  & \begin{tabular}[c]{@{}l@{}}If there is no hotel availability , I will \\ accept a guesthouse. Is one availabel ?\end{tabular}                                     & \begin{tabular}[c]{@{}l@{}}Hotel-Inform.Type: guesthouse\\ {\color{red}Hotel-Inform.Stars: 4}\end{tabular}                                                                                                  & Hotel-Inform.Type: guesthouse                                                                                                                                        \\ \cline{2-5}

                                                                              & MUL0666.json  & \begin{tabular}[c]{@{}l@{}}Just please book for that room for 2\\ nights .\end{tabular}                                                                           & \begin{tabular}[c]{@{}l@{}}{\color{red}Hotel-Inform.Price: cheap}\\ Hotel-Inform.Stay: 2\end{tabular}                                                                                                       & Hotel-Inform.Stay: 2                                                                                                                                                 \\ \hline

\end{tabular}
}

\caption{Example of different error types of dialogue acts. The red color in the table highlights incorrect annotations and corresponding repaired results. Note that MultiWOZ 2.2 is excluded from the table because it added missing dialogue act annotations and the remainings are the same as MultiWOZ 2.1.}
\label{tab:da_example}
\end{table*}

To address the limitations above, we introduce an updated version, MultiWOZ 2.3\footnote[2]{\href{https://github.com/lexmen318/MultiWOZ-coref}{https://github.com/lexmen318/MultiWOZ-coref}}. Our contributions are as follow:

\begin{itemize}
\item We differentiate incorrect annotations in dialogue acts from those in dialogue states, and unify annotations of dialogue acts and dialogue states to ensure their consistency when publishing the updated dataset, MultiWOZ 2.3. 
\item We introduce co-reference features to annotations of the dialogue dataset to enhance the performances of dialogue systems in the new version.

\item We re-benchmark a few SOTA models for dialogue state tracking  (DST) and natural language understanding (NLU) tasks and provide a fair comparison using the updated dataset.
\end{itemize}

\section{Annotation Corrections}

The inconsistent annotations in the MultiWOZ dataset were caused by disparate interpretations from involved annotators during a crowdsourcing process. These errors can occur even when annotators attempt to apply unified rules.  After analyzing annotation errors in both dialogue acts and dialogue states, we perform the following two data corrections.

\begin{table*}[!htb]
\centering
\resizebox{\textwidth}{32mm}{
\begin{tabular}{|l|l|l|l|} \hline

\textbf{Dialogue ID}  & \textbf{Utterance}  & \textbf{MultiWOZ 2.1}     & \textbf{MultiWOZ 2.3}                                                                              \\ \hline

\multirow{2}{14mm}{MUL2602.json} & \begin{tabular}[c]{@{}l@{}}\textit{User}: Can you recommend me a nightclub where I can get jiggy with\\ it?\\ \textit{Sys}: Well, I think the jiggiest nightclub in town is the Soul Tree\\ Nightclub, right in centre city! Plis the entrance fee isonly 4\\ pounds\end{tabular} & \begin{tabular}[c]{@{}l@{}}{\color{red}a-type=night club}\\ a-name=not mentioned\\ a-area=not mentioned\end{tabular}              & \begin{tabular}[c]{@{}l@{}}{\color{red}a-type=nightclub}\\ a-name=not mentioned\\ a-area=not mentioned\end{tabular}               \\ \cline{2-4}

                              & \begin{tabular}[c]{@{}l@{}}\textit{User}: That is perfect can I have the postcode please?\\ \textit{Sys}: Sure! The postcode is cb23qf\end{tabular}                                                                                                                               & \begin{tabular}[c]{@{}l@{}}{\color{red}a-type=night club}\\ {\color{red}a-name=not mentioned}\\ a-area=not mentioned\end{tabular}              & \begin{tabular}[c]{@{}l@{}}{\color{red}a-type=nightclub}\\ {\color{red}a-name=soul tree nightclub}\\ a-area=not mentioned\end{tabular}         \\ \hline

\multirow{2}*{MUL1455.json} & \begin{tabular}[c]{@{}l@{}}\textit{User}: I am also looking for a moderately priced chinses restaurant\\ located in the north\\ \textit{Sys}: Golden wok is the moderate price range and in the north\\ area would you like me to book it for you?\end{tabular}                   & \begin{tabular}[c]{@{}l@{}}r-food=chinese\\ r-pricerange=moderate\\ r-name=not mentioned\\ r-area=north\end{tabular} & \begin{tabular}[c]{@{}l@{}}r-food=chinese\\ r-pricerange=moderate\\ r-name=not mentioned\\ r-area=north\end{tabular} \\ \cline{2-4}

                              & \begin{tabular}[c]{@{}l@{}}\textit{User}: Can I get the address and phone number please?\\ \textit{Sys}: Of course - the address is 191 Histon Road Chesterton cb43hl\\ and the phone number is 01223350688\end{tabular}                                                          & \begin{tabular}[c]{@{}l@{}}r-food=chinese\\ r-pricerange=moderate\\ {\color{red}r-name=not mentioned}\\ r-area=north\end{tabular} & \begin{tabular}[c]{@{}l@{}}r-food=chinese\\ r-pricerange=moderate\\ {\color{red}r-name=golden wok}\\ r-area=north\end{tabular}    \\ \hline
\end{tabular}
}

\caption{Example of updates on dialogue states. The red color in the figure highlights incorrect dialogue states and corresponding updated results. Note that MultiWOZ 2.2 is excluded from the figure because it is the same to MultiWOZ 2.1 in terms of inconsistent tracking. ``a" and ``r" used as slot names in the right two columns are abbreviations for ``attraction" and ``restaurant" respectively.}
\label{tab:inconsist}
\end{table*}

\subsection{Dialogue Act Corrections}

The annotations for user dialogue acts were originally introduced by~\citet{lee_etal_2019_convlab}. Following the pipeline provided in ConvLab, \citet{eric_etal_2020_multiwoz} re-annotated dialogue acts for both systems and users in MultiWOZ 2.1. We broadly categorize the incorrect annotations into three types (Table~\ref{tab:da_example}) based on our observations:


\begin{itemize}
\item \textbf{Under-annotated}: Annotation errors under this category are due to insufficient annotation even when exact information is available in the given dialogue utterances. The missing annotations should be added to the corresponding slots. 
\item \textbf{Over-annotated}: Sometimes, incorrect annotations are put down even when no corresponding information can be identified in the utterances.  The over-annotated values should be removed to avoid confusion. 
\item \textbf{Wrongly-annotated}: This category refers to slots with incorrect values (or span information) and should be fixed.

\end{itemize}

We apply two rules to sequentially correct ``dialog\_act" annotations: a) we use customized filters to select credible predictions generated from a MultiWOZ 2.1 pre-trained BERTNLU model ~\citep{zhu_etal_2020_convlab} and merge them with original ``dialog\_act" annotations; b) we use assorted regular expressions to further clean ``dialog\_act" annotations from the previous step.

To fairly evaluate the quality of modified annotations, we sampled 100 dialogues from the test set and manually re-annotated the dialogue acts. Table~\ref{da-correctness} exhibits the ratios of ``dialog\_act" annotations of different datasets in terms of slot level and turn level using the manually-annotated 100 dialogues as golden annotations.

\begin{table}[!htb]

\renewcommand{\tabularxcolumn}[1]{m{#1}}
\newcolumntype{Y}{>{\centering\arraybackslash}X}

\begin{center}
    \begin{tabularx}{\columnwidth}{|Y|Y|Y|Y|}
        \hline
        \textbf{Version} & \textbf{Rule} & \textbf{Slot Level} & \textbf{Turn Level} \\
        \hline
        \textbf{2.1/2.2} & \begin{tabular}{c}
             Strict \\
             Relax* \\
        \end{tabular} & \begin{tabular}{c}
             77.59\%  \\
             82.94\%  \\
        \end{tabular} & \begin{tabular}{c}
             68.83\%  \\
             77.19\%  \\
        \end{tabular} \\
        \hline
        \textbf{2.3} & \begin{tabular}{c}
             Strict \\
             Relax* \\
        \end{tabular} & \begin{tabular}{c}
             84.12\%  \\
             90.74\%  \\
        \end{tabular} & \begin{tabular}{c}
             76.09\%  \\
             86.83\%  \\
        \end{tabular} \\
        \hline

    \end{tabularx}
    \caption{\label{da-correctness} A comparison of annotation correctness ratios of ``dialog\_act"  for MultiWOZ 2.1/2.2 and coref. The ``Relax" rule indicates that the values of insignificant slots like ``general-xxx" and ``none" are removed.}
\end{center}
\end{table}

\begin{figure*}[!htb]

 \center

  \includegraphics[width=\textwidth]{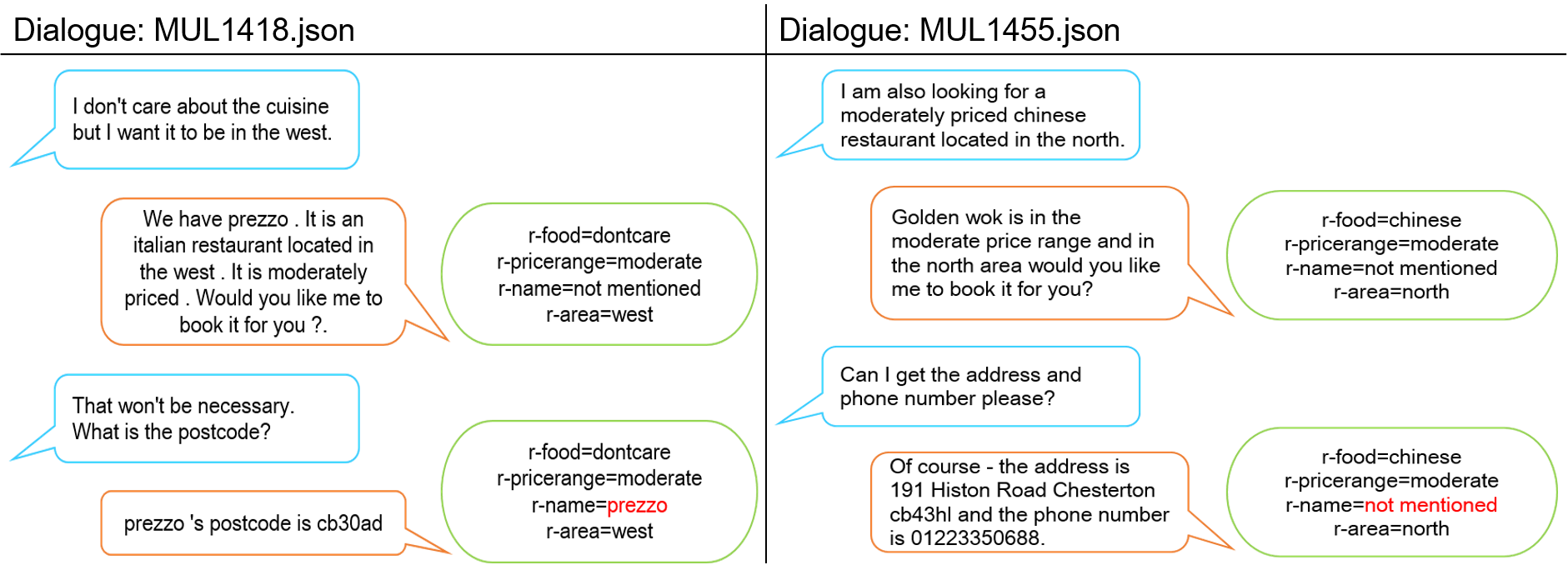}

  \caption{Examples of inconsistent tracking on dialogue states of two different dialogues in similar scenarios from MultiWOZ 2.1. In the left column, dialogue \textit{MUL1418.json} updates slot \textit{r-name} with ``prezzo" recommended by the system. However, for dialogue \textit{MUL1455.json} in the right column, the value of slot \textit{r-name} is remained as ``not mentioned" even though ``golden wok" is recommended by the system. ``r" in the light green rectangle is an abbreviation for ``restaurant".}

  \label{fig:inconsist3}

\end{figure*}

We added 24,405 slots and removed 4,061 slots in the ``dialog\_act" annotations. Roughly 16,800 slots are modified according to our estimation. Also note that in Table~\ref{tab:da_example}, boundaries for the three types are not strictly drawn. \textit{PMUL2596.json} under wrongly-annotated type can also be treated as an under-annotated error when slot \textit{Taxi-Inform.Dest} is missing. 

Adding and removing operations for ``dialog\_act" annotations cause mismatches in paired span indices. When aligning span information with the modified dialogue acts, we note that original span information also contains incorrect annotations, such as abnormal span with ending index ahead of the starting index, incorrect span, and drifted span. The errors are all corrected, along with those for dialogue acts.

\subsection{Dialogue State Corrections}

The fixed ``dialog\_act" and the ``span\_info" annotations are propagated into the dialogue state annotations(i.e., `` metadata" annotations), because we need to maintain the consistency among them.  

Since the repairing for dialogue states is based on cleaned dialogue acts, we use the following rules to guide updating dialogue state annotations (Table~\ref{tab:inconsist}): 

\begin{itemize}
\item \textbf{Slot Value Normalization}: Multiple slots values exist in MultiWOZ 2.2 due to a mismatch between given utterances and ontology, for example, ``16:00" and ``4 PM". This potentially leads to incomplete matching, as the values are not normalized. To this end, we follow the way how MultiWOZ 2.1 normalizes slot values based on utterances. 

\item \textbf{Consistent Tracking Strategy}: The inconsistent tracking strategy (Figure~\ref{fig:inconsist3}) was initially discussed (but not solved) in MulitWOZ 2.2. We track the user's requirements from slot values informed by the user, recommended by the system, and implicitly agreed by the user. We apply two sub-rules to resolve the implicit agreements: a) if an informing action is from the user to the system, the informed values are propagated to the next turn of dialogue states; b) if an informing/recommending action is from the system to the user, the informed or recommended values are propagated to the next turn of dialogues states if and only if one item is included. Multiple items are not considered to be valid in the implicit agreement settings.

\end{itemize}

\begin{table}[!htb]
\begin{center}
    \begin{tabular}{|c|c|c|}
        \hline
        \textbf{Fixing Type} & \textbf{Count} & \textbf{Ratio} \\
        \hline
        No Change & 2,476,666 & 98.68\% \\
        \hline
        Value Filled & 20,639 & 0.82\% \\
        \hline
        Value Changed & 11,649 & 0.46\% \\
        \hline
        Value Removed & 221 & 0.01\% \\
        \hline
        Value dontcare & 563 & 0.02\% \\
        \hline
    \end{tabular}
    \caption{Percentage of slots' values changed in MultiWOZ 2.3 and MultiWOZ2.1, respectively, for ``metadata" annotations. ``Value Filled" stands for a value-filled from null, ``none" or ``not mentioned". ``Value Removed" means a slot value is changed to ``not mentioned" or null. ``Value dontcare" stands for slot values filled with ``dontcare".}
    \label{ds-correctness}
\end{center}
\end{table}

Table~\ref{ds-correctness} shows statistics on the type of corrections we have made on the ``metadata" annotations. Note that ``dontcare" value is singled out during repairing since it is a significant factor (Table \ref{tab:slot-gate}) on slot gate classifications in the TRADE model~\citep{wu_etal_2019_transferable}.

\begin{table*}[ht]

\begin{center}
    \begin{tabularx}{\textwidth}{|c|c|}
    \hline
    \textbf{Dialogue ID} & \textbf{Utterances}\\
    \hline
    PMUL1815.json & \begin{tabular}{X}
    I'm traveling to Cambridge from london liverpool street arriving by 11:45 \underline{the day} \textcolor{orange}{{\textit{(saturday)}}} of my hotel booking. \\      
    \end{tabular} \\
    \hline
    
    PMUL2049.json & \begin{tabular}{X}
    Thank you, can you also help me find a restaurant that is in \underline{the same area} \textcolor{orange}{\textit{(centre)}} as the Parkside pools?
    \end{tabular} \\
    \hline
    
    PMUL2512.json & \begin{tabular}{X}
    Thanks! I'm going to hanging out at \underline{the college} \textcolor{orange}{\textit{(christ college)}} late tonight, could you get me a taxi back to \underline{the hotel} \textcolor{orange}{\textit{(the express by holiday inn cambridge)}} at 2:45?
    \end{tabular} \\
    \hline
    
    \end{tabularx}
    \caption{Examples of co-reference annotations. Co-reference values are added to the original utterances and marked as light orange italic inside the brackets.} 
    \label{tab:coreference}
\end{center}
\end{table*}

\section{Enhance Dataset with Co-referencing}


MultiWOZ contains a considerable amount of co-reference and ellipsis. As shown in Tabel~\ref{tab:coreference}, co-referencing frequently occurs in the cross-domain dialogues, especially when aligning the value of ``Name" slot from a hotel (or restaurant) domain with those of ``Departure/Destination" slots for taxi/train domains. The lack of co-reference annotations leads to poor performances presented in existing DST models.  


A number of task-oriented dialogue models leveraged datasets enhanced with co-referencing features to achieve SOTA results~\citep{ferreira2020coreference}. By including co-reference in CamRest676~\citep{wen_etal_2017_network}, GECOR~\citep{quan_etal_2019_gecor} showed significant performance improvement compared to the baseline models. Through restoring incomplete utterances by annotating the dataset with co-reference labels, ~\citet{pan_etal_2019_improving} boosted response quality of dialogue systems. \citet{su_etal_2019_improving} re-wrote utterances to cover co-referred and omitted information to realize notable success on their proposed model. 

In MultiWOZ 2.1, the distributions of co-referencing among different slots are presented in Table~\ref{tab:coref-stat}. In total, 20.16\% dialogues are annotated with co-reference in the dataset, indicating the importance of co-referencing annotation.

\subsection{Annotation for Co-reference in Dialogue}

\begin{figure}[ht]

 \center

  \includegraphics[width=\linewidth]{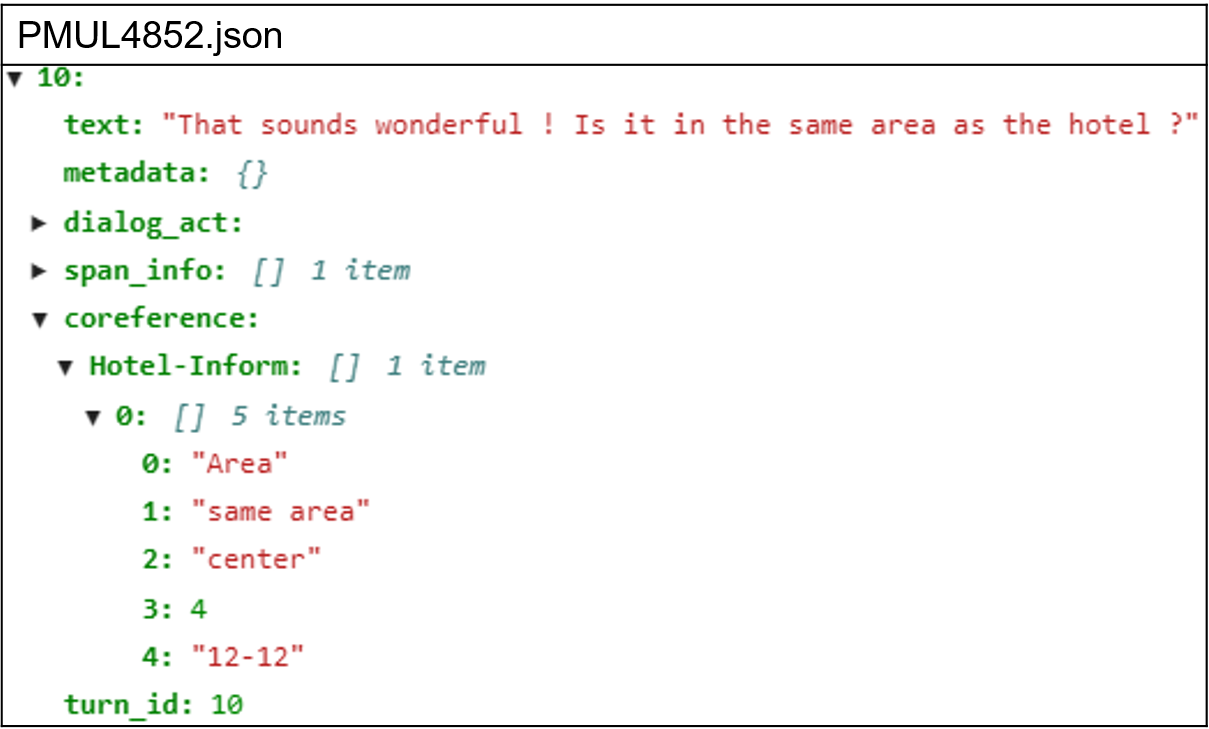}

  \caption{Example of a co-referencing annotation. If the current turn involves more than one co-referencing relationships, all annotations will be gathered under the ``coreference" key. The number ``10" at the top left corner indicates the ``turn\_id" of dialogue \textit{PMUL4852.json}.}

  \label{fig:coref}

\end{figure}

The ``coreference" annotations are applied to all ``dialog\_act" slots having co-referencing relationships with other slots. The annotation takes a  ``Domain-Intent" format, including five parts: slot name, slot value in the current turn, referred value, referred turn id, and spans of referred value in the referred turn. Figure~\ref{fig:coref} depicts an example of ``coreference" annotation and the corresponding values for the five parts are ``Area", ``same area", ``center", ``4", ``12-12" under ``Hotel-Inform"

We apply co-referencing annotations to problematic slots when necessary, for example, ``Area/Price/People/Day/Depart/Dest/Arrive". The co-referencing annotations are added sequentially:

\begin{itemize}
\item We use first regular expressions to locate co-reference slots;
\item Based on the current dialogue states, we trace back to the history utterances where the co-referred slots are first encountered; 
\item We use the corresponding dialogue acts with paired span information to retrieve co-referred values.
\end{itemize}

In total, we added 3,340 co-referencing annotations for ``dialog\_act". 

\begin{table}[!htb]
\begin{center}
    \begin{tabular}{|c|c|c|}
        \hline
        \textbf{Slot} & \textbf{Count} & \textbf{Ratio} \\
        \hline
        Taxi.Depart & 844 & 24.82\% \\
        \hline
        H/R/A.Area & 786 & 23.12\% \\
        \hline
        Taxi.Dest & 706 & 20.76\% \\
        \hline
        H/R/A/T.Day & 409 & 12.03\% \\
        \hline
        H/R.Price & 354 & 10.41\% \\
        \hline
        H/R/T.People & 201 & 5.91\% \\
        \hline
        Taxi.Arrive & 92 & 2.71\% \\
        \hline
    \end{tabular}
    \caption{Statistics of co-reference annotations. H/R/A/T represent ``Hotel", ``Restaurant", ``Attraction" and ``Train", respectively.}
    \label{tab:coref-stat}
\end{center}
\end{table}

Table~\ref{tab:coref-stat} shows the statistics of the amount of ``coreference" annotations for each slot type. We can see the most common co-referencing relationship is from ``Taxi-Dest/Depart" and ``xxx-Area", followed by ``Day", ``Price", ``People" and ``Arrive".

\begin{table*}[!htb]
\centering
\resizebox{\textwidth}{35mm}{
\begin{tabular}{|l|l|l|l|}
\hline
\multicolumn{1}{|c|}{\textbf{Dialogue ID}} & \multicolumn{1}{c|}{\textbf{Goal description}}  & \multicolumn{1}{c|}{\textbf{Original annotation}} & \multicolumn{1}{c|}{\textbf{New annotation}}                                                                                                                                                                                                                                               \\ \hline

PMUL4372.json                     & \begin{tabular}[c]{@{}l@{}}You are slo looking for a \textit{place to stay}.\\ The hotel should \textit{include free parking} and\\ should be in the {\color{red}\textit{same price range as the}}\\ {\color{red}\textit{restaurant}}.\\ The hotel should \textit{include free wifi}.\\ Once you find the \textit{hotel}, you want to book it for\\ {\color{red}\textit{the same group of people}} and {\color{red}\textit{3 nights}} starting\\ from {\color{red}\textit{the same day}}.\\ If the booking fails how about {\color{red}\textit{1 nights}}.\\ Make sure you get the {\color{red}\textit{reference number}}.\end{tabular} & \begin{tabular}[c]{@{}l@{}}\textbf{Constraint}\\ hotel.parking=yes\\ {\color{red}hotel.pricerange=expensive}\\ hotel.internet=yes\\ {\color{red}hotel.people=3}\\ {\color{red}hotel.day=wednesday}\\ {\color{red}hotel.stay=3}\end{tabular} & \begin{tabular}[c]{@{}l@{}}\textbf{Constraint}\\ hotel.parking=yes\\ {\color{red}hotel.pricerange={[}restaurant,}\\ {\color{red}pricerange{]}}\\ hotel.internet=yes\\ {\color{red}hotel.people={[}restaurant, people{]}}\\ {\color{red}hotel.day={[}restaurant, day{]}}\\ {\color{red}hotel.stay=3$\rightarrow$1}\\ \\ \textbf{Request}\\ {\color{red}hotel.Ref=?}\end{tabular} \\ \hline

PMUL2512.json                     & \begin{tabular}[c]{@{}l@{}}You also want to book a \textit{taxi} to {\color{red}\textit{commute}}\\{\color{red}\textit{ between the two places}}.\\ You want to leave the \textit{attraction} by \textit{02:45}.\\ Make sure you get \textit{contact number} and \textit{car type}.\end{tabular}                                                                                                                                                                                                                                   & \begin{tabular}[c]{@{}l@{}}\textbf{Constraint}\\ taxi.leaveAt=02:45\\ \\ \textbf{Request}\\ taxi.phone=?\\ taxi.car type=?\end{tabular}                                                          & \begin{tabular}[c]{@{}l@{}}\textbf{Constraint}\\ {\color{red}taxi.departure={[}attraction, None{]}}\\ {\color{red}taxi.destination={[}hotel, None{]}}\\ taxi.leaveAt=02:45\\ \\ \textbf{Request}\\ taxi.phone=?\\ taxi.car type=?\end{tabular}                                                                                \\ \hline

\end{tabular}
}
\caption{Examples of co-reference annotations in the user goal. The red color highlights the difference between the original and new annotations}
\label{goal-example}
\end{table*}

\subsection{Annotation for Co-reference in User Goal}



During the data collection process, the user converses with the system, following a given goal description~\citep{budzianowski2018large}. Co-reference in the user utterances is derived from co-reference in user goals. However, the goal annotation, represented as several constraints and requests, is not consistent with the goal description and does not implement co-reference features. Table \ref{goal-example} shows two examples of user goals with co-reference. 
The original goal annotation misses a request, three constraints and all co-reference relations.
The right arrow (hotel.stay=3$\rightarrow$1) indicates a possible goal change during a dialogue.
The co-referencing relations are represented as referenced domains and slots. Note that the referenced slot of ``taxi.departure/taxi.destination" is uncertain because the departure may be a name, an address, or ``the attraction". \textit{PMUL2512.json} in Table~\ref{tab:coreference} shows the relation between the goal and utterance: the co-reference annotations of ``the college" and ``the hotel" realize the the referenced slot of ``taxi.departure/taxi.destination" in the new annotations of user goal.

To introduce co-referencing annotation into user goals, we use regular expressions to extract all slot-value pairs and co-referencing relations from the goal descriptions. We manually check 150 random samples and confirm the correctness of the new goal annotations. The new goal annotations may contribute to better user simulators\citep{schatzmann2007agenda,gur2018user}, which generate user responses or evaluate system performances based on user goals.

\section{Benchmarks and Experimental Results}
The updated dataset is evaluated for a natural language understanding task and a DST task. Experiment results are produced to re-benchmark a few SOTA models.

\begin{table*}[t]
\centering
\begin{tabular}{|c|c|c|}
    \hline
    \textbf{Dataset} & \textbf{F1(Slot/Intent/Both)} & \textbf{Utter. Acc.(Slot/Intent/Both)} \\
    \hline
    MultiWOZ 2.1 & \begin{tabular}{ccc}
         81.18/88.34/83.77
    \end{tabular} & \begin{tabular}{ccc}
         81.89/86.23/71.68
    \end{tabular} \\
    \hline
    MultiWOZ 2.2 & \begin{tabular}{ccc}
        80.61/88.34/83.41
    \end{tabular} & \begin{tabular}{ccc}
         81.94/86.41/71.85
    \end{tabular} \\
    \hline
    MultiWOZ 2.3 & \begin{tabular}{ccc}
        \textbf{89.03}/\textbf{90.73}/\textbf{89.65}
    \end{tabular}& \begin{tabular}{ccc}
         \textbf{87.33}/\textbf{88.56}/\textbf{78.33}
    \end{tabular} \\
    \hline
\end{tabular}
\caption{\label{tab:nlu-benchmark} Performance of BERTNLU on different datasets based on F1 score and utterance accuracy for slots, intents and both, respectively. Utterance accuracy is defined as the average accuracy of predicting all the slots, intents or both in an utterance correctly.}
\end{table*}

\begin{table*}[t]
    \centering
    \begin{tabular}{|c|c|c|c|}
    \hline
    \textbf{Dataset} & \textbf{Pointer(P/R/F1)} & \textbf{Dontcare(P/R/F1)} & \textbf{None(P/R/F1)}  \\
    \hline
    MultiWOZ 2.1 & 94.97/93.75/94.35 & 58.73/32.51/41.85 & 98.25/98.82/98.53 \\
    \hline
    MultiWOZ 2.2 & 94.22/94.42/94.32 & 60.21/34.60/43.91 & 98.42/98.64/98.53 \\
    \hline
    MultiWOZ 2.3 & \textbf{96.41/96.15/96.28} & \textbf{67.80/41.62/51.58} & \textbf{98.79/99.11/98.95} \\
    \hline
    \end{tabular}
    \caption{Classification on slot gate for TRADE using different datasets. ``Pointer", ``dontcare" and ``none" are three different slot gate classes. Precision, recall, and F1-score are used as metrics to evaluate among all datasets.}
    \label{tab:slot-gate}
\end{table*}

\subsection{Dialogue Actions with Natural Language Understanding Benchmarks}

BERTNLU~\citep{zhu_etal_2020_convlab} is introduced for dialogue natural language understanding. It tops extra two multilayer perceptron (MLP) layers on BERT~\citep{devlin_etal_2019_bert} for slot recognition and intent classification~\citep{2019arXiv190210909C}, respectively. In practice, BERTNLU achieves better performance on classification and tagging tasks by including historical context and finetuning all parameters. We implement BERTNLU with inputs of current utterance plus the previous three history turns and finetune it based on the dialogue act annotations. The model's performance is evaluated by calculating F1 scores for intents, slots, and for both. Additionally, we use utterance accuracy as another metric to assess the model's effectiveness in understanding what the user expresses in an utterance. We score each utterance either 0 or 1 according to whether the predictions of all the slots, intents, or both in an utterance match the correct labels. The utterance accuracy is characterized as the average of this score across all utterances. Table \ref{tab:nlu-benchmark} shows the performance of BERTNLU on different datasets (including dialogue utterances from both user and system sides) based on the above evaluation metrics.

\subsection{Dialogue State Tracking Benchmarks\footnote[3]{Full benchmarks with various models are available in Appendix~\ref{sec:appendix2}}}

Multiple neural network-based models have been proposed to improve joint goal accuracy of dialogue state tracking tasks. Existing belief state trackers could be roughly divided into two classes: span-based and candidate-based. The former approach~\citep{zhang2019find, heck_etal_2020_trippy, lee_etal_2019_sumbt} directly extracts slot values from dialogue history, while the latter approach ~\citep{wu_etal_2019_transferable} is to perform classification on candidate values, assuming all candidate values are included in the predefined ontology. To evaluate our updated dataset for DST task, we run experiments on TRADE~\citep{wu_etal_2019_transferable} and SUMBT~\citep{lee_etal_2019_sumbt}. 

SUMBT uses a multi-head attention mechanism to capture relations between domain-slot types and slot values presented in the utterances. The attended context words are collected as slot values for corresponding slots. TRADE uses a pointer to differentiate, for a particular domain-slot, whether the slot value is from the given utterance or the predefined vocabulary. Both models perform predictions slot by slot and treat all slots equally.


Following the convention in dialogue state tracking task, joint goal accuracy is used to evaluate the models' performances for different datasets. The models also experiment with co-referencing enhanced datasets. Table \ref{tab:DST} summarizes the joint goal accuracy of the two models using different datasets. 

\begin{table}[!htb]
\begin{center}
    \begin{tabular}{|c|c|c|}
        \hline
        \textbf{Dataset} & \textbf{SUMBT} & \textbf{TRADE} \\
        \hline
        MultiWOZ 2.0 & 46.6\%$^\blacklozenge$ & 48.6\%$^\blacklozenge$ \\

        MultiWOZ 2.1 & 49.2\% & 45.6\% \\

        MultiWOZ 2.2 & 49.7\% & 46.6\% \\
        \hline
        MultiWOZ 2.3 & \textbf{52.9}\% & \textbf{49.2}\% \\

        MultiWOZ-coref & \textbf{54.6}\% & \textbf{49.9}\% \\
        \hline
    \end{tabular}
    \caption{Joint goal accuracy of SUMBT and TRADE over different versions of dataset. MultiWOZ-coref refers to the dataset with co-reference applied. $\blacklozenge$ means the accuracy scores are adopted from the published papers.}
    \label{tab:DST}
\end{center}
\end{table}

\subsection{Experimental Analysis}


As shown in Tables \ref{tab:nlu-benchmark} and \ref{tab:DST}, substantial performance increases are achieved with the enhanced datasets compared to the previous datasets. BERTNLU trained using our dataset outperforms others with a margin of 5\% improvement on both metric of F1-score and utterance accuracy. In the task of DST, models trained using our datasets also show superiority to those trained with the previous version MultiWOZ.  By applying co-referencing features to dialogue state tracking, the joint goal accuracy is improved to approximately 55\% using SUMBT.

\section{Discussion}
\begin{figure}[!htb]

 \center

  \includegraphics[width=\columnwidth]{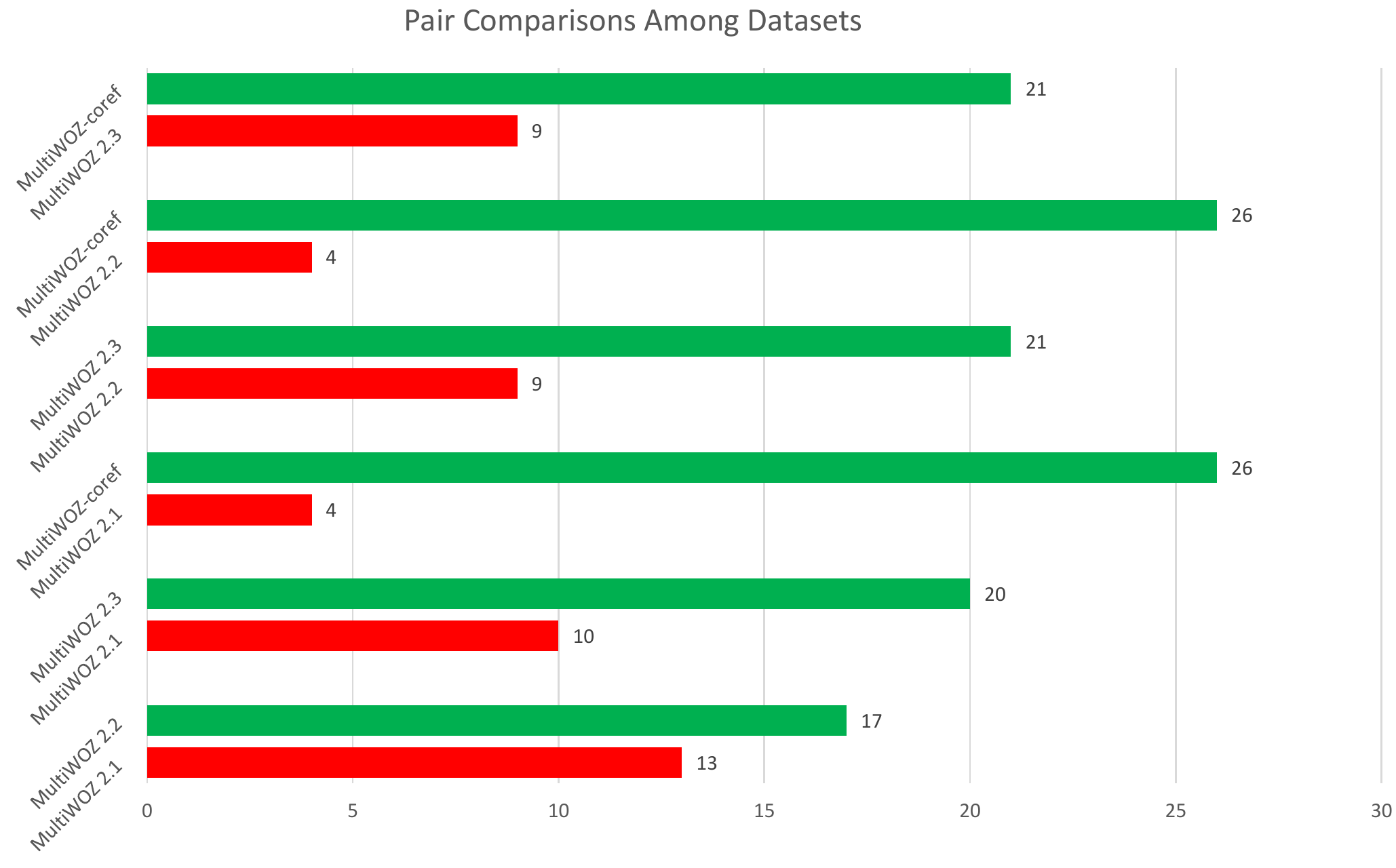}

  \caption{Pairwise comparison between two datasets in terms of the number of higher accuracy slots. In total, there are 30 valid slots in the DST task. The number on top of each bar indicates the number of winning slots in comparison.}

  \label{fig:paircom}

\end{figure}

Note that SUMBT initially focused on MultiWOZ 2.0. Fixing dialogue states leads to enhanced data quality in MultiWOZ 2.1. This study adopts a rule-based method to correct the identified errors in MultiWOZ 2.1 further. With a customized pre-process\footnote[4]{Scores shown in Table~\ref{tab:DST} are achieved by using pre-process scripts provided by SUMBT and TRADE.} script for SUMBT, the joint goal accuracy can reach 54.54\% for MultiWOZ 2.3 and 56.09\% for MultiWOZ-coref, respectively. Since multiple slot values are allowed for MultiWOZ 2.2, it is not practical to identify errors in the dialogue states. We do not base this study on MultiWOZ 2.2 at this stage. Figure~\ref{fig:paircom} shows pairwise comparisons between two datasets on the benchmarked scores. Our dataset (MultiWOZ 2.3) tops all the scores compared to previously updated datasets in all MultiWOZ specified slots. Details of slot accuracies are presented in Table~\ref{tab:sumbt-merge} at Appendix~\ref{sec:appendix1}.  As shown in Table~\ref{tab:sumbt-merge}, our dataset achieves the best performance for 17 out of all 30 slots. The performance is further enhanced with the co-reference version (24 out of all 30).

Table~\ref{tab:slot-gate} shows precision, recall, and F1-score of slot gate classifications in the TRADE model across different datasets. For the three different classes, our dataset achieves top performances. As a result of the carefully designed error correction (Table~\ref{DAtoState} in Appendix~\ref{sec:appendix}), our dataset outperforms others by at least 9\% in all metrics for the ``dontcare" gate. 


Based on the contexts presented in utterances, we have fixed the dialogue acts and removed the inconsistency between dialogue acts and states. Span indices in the dialogue acts are further fixed with co-reference information introduced. By closely aligning the annotations to corresponding utterances mentioned above, we remove the inconsistency introduced by annotating a  Wizard-Of-Oz dataset.

\section{Conclusion}
MultiWOZ datasets (2.0-2.2) are widely used in dialogue state tracking and other dialogue related subtasks. Mainly based on MultiWOZ 2.1, we publish a refined version, named MultiWOZ 2.3. After correcting annotations for dialogue acts and dialogue states, we introduce co-reference annotations, which supports future research to consider discourse analysis in building task-oriented dialogue systems. We re-benchmark the refined dataset using some competitive models. The experimental results show significant improvements for the associated scores, verifying the utility of this dataset. We hope to attract more alike research works to improve the quality of MultiWOZ datasets further.

\bibliography{anthology,eacl2021}

\begin{thebibliography}{31}
\expandafter\ifx\csname natexlab\endcsname\relax\def\natexlab#1{#1}\fi

\bibitem[{Budzianowski et~al.(2018)Budzianowski, Wen, Tseng, Casanueva, Ultes,
  Ramadan, and Ga{\v{s}}i{\'c}}]{budzianowski2018large}
Pawe{\l} Budzianowski, Tsung-Hsien Wen, Bo-Hsiang Tseng, I{\~n}igo Casanueva,
  Stefan Ultes, Osman Ramadan, and Milica Ga{\v{s}}i{\'c}. 2018.
\newblock \href {https://doi.org/10.18653/v1/D18-1547} {{M}ulti{WOZ} - a
  large-scale multi-domain wizard-of-{O}z dataset for task-oriented dialogue
  modelling}.
\newblock In \emph{Proceedings of the 2018 Conference on Empirical Methods in
  Natural Language Processing}, pages 5016--5026, Brussels, Belgium.
  Association for Computational Linguistics.

\bibitem[{{Chen} et~al.(2019){Chen}, {Zhuo}, and {Wang}}]{2019arXiv190210909C}
Qian {Chen}, Zhu {Zhuo}, and Wen {Wang}. 2019.
\newblock \href {http://arxiv.org/abs/1902.10909} {{BERT for Joint Intent
  Classification and Slot Filling}}.
\newblock \emph{arXiv e-prints}, page arXiv:1902.10909.

\bibitem[{Chen et~al.(2019)Chen, Chen, Qin, Yan, and
  Wang}]{chen_etal_2019_semantically}
Wenhu Chen, Jianshu Chen, Pengda Qin, Xifeng Yan, and William~Yang Wang. 2019.
\newblock \href {https://doi.org/10.18653/v1/P19-1360} {Semantically
  conditioned dialog response generation via hierarchical disentangled
  self-attention}.
\newblock In \emph{Proceedings of the 57th Annual Meeting of the Association
  for Computational Linguistics}, pages 3696--3709, Florence, Italy.
  Association for Computational Linguistics.

\bibitem[{Devlin et~al.(2019)Devlin, Chang, Lee, and
  Toutanova}]{devlin_etal_2019_bert}
Jacob Devlin, Ming-Wei Chang, Kenton Lee, and Kristina Toutanova. 2019.
\newblock \href {https://doi.org/10.18653/v1/N19-1423} {{BERT}: Pre-training of
  deep bidirectional transformers for language understanding}.
\newblock In \emph{Proceedings of the 2019 Conference of the North {A}merican
  Chapter of the Association for Computational Linguistics: Human Language
  Technologies, Volume 1 (Long and Short Papers)}, pages 4171--4186,
  Minneapolis, Minnesota. Association for Computational Linguistics.

\bibitem[{Eric et~al.(2020)Eric, Goel, Paul, Sethi, Agarwal, Gao, Kumar, Goyal,
  Ku, and Hakkani-Tur}]{eric_etal_2020_multiwoz}
Mihail Eric, Rahul Goel, Shachi Paul, Abhishek Sethi, Sanchit Agarwal, Shuyang
  Gao, Adarsh Kumar, Anuj Goyal, Peter Ku, and Dilek Hakkani-Tur. 2020.
\newblock \href {https://www.aclweb.org/anthology/2020.lrec-1.53} {{M}ulti{WOZ}
  2.1: A consolidated multi-domain dialogue dataset with state corrections and
  state tracking baselines}.
\newblock In \emph{Proceedings of The 12th Language Resources and Evaluation
  Conference}, pages 422--428, Marseille, France. European Language Resources
  Association.

\bibitem[{Ferreira~Cruz et~al.(2020)Ferreira~Cruz, Rocha, and
  Lopes~Cardoso}]{ferreira2020coreference}
Andr{\'e} Ferreira~Cruz, Gil Rocha, and Henrique Lopes~Cardoso. 2020.
\newblock Coreference resolution: Toward end-to-end and cross-lingual systems.
\newblock \emph{Information}, 11(2):74.

\bibitem[{Gao et~al.(2019)Gao, Sethi, Agarwal, Chung, and
  Hakkani-Tur}]{gao_etal_2019_dialog}
Shuyang Gao, Abhishek Sethi, Sanchit Agarwal, Tagyoung Chung, and Dilek
  Hakkani-Tur. 2019.
\newblock \href {https://doi.org/10.18653/v1/W19-5932} {Dialog state tracking:
  A neural reading comprehension approach}.
\newblock In \emph{Proceedings of the 20th Annual SIGdial Meeting on Discourse
  and Dialogue}, pages 264--273, Stockholm, Sweden. Association for
  Computational Linguistics.

\bibitem[{G{\"u}r et~al.(2018)G{\"u}r, Hakkani-T{\"u}r, T{\"u}r, and
  Shah}]{gur2018user}
Izzeddin G{\"u}r, Dilek Hakkani-T{\"u}r, Gokhan T{\"u}r, and Pararth Shah.
  2018.
\newblock User modeling for task oriented dialogues.
\newblock In \emph{2018 IEEE Spoken Language Technology Workshop (SLT)}, pages
  900--906. IEEE.

\bibitem[{Heck et~al.(2020)Heck, van Niekerk, Lubis, Geishauser, Lin, Moresi,
  and Gasic}]{heck_etal_2020_trippy}
Michael Heck, Carel van Niekerk, Nurul Lubis, Christian Geishauser, Hsien-Chin
  Lin, Marco Moresi, and Milica Gasic. 2020.
\newblock \href {https://www.aclweb.org/anthology/2020.sigdial-1.4}
  {{T}rip{P}y: A triple copy strategy for value independent neural dialog state
  tracking}.
\newblock In \emph{Proceedings of the 21th Annual Meeting of the Special
  Interest Group on Discourse and Dialogue}, pages 35--44, 1st virtual meeting.
  Association for Computational Linguistics.

\bibitem[{Henderson et~al.(2014)Henderson, Thomson, and
  Williams}]{henderson_etal_2014_second}
Matthew Henderson, Blaise Thomson, and Jason~D. Williams. 2014.
\newblock \href {https://doi.org/10.3115/v1/W14-4337} {The second dialog state
  tracking challenge}.
\newblock In \emph{Proceedings of the 15th Annual Meeting of the Special
  Interest Group on Discourse and Dialogue ({SIGDIAL})}, pages 263--272,
  Philadelphia, PA, U.S.A. Association for Computational Linguistics.

\bibitem[{Kim et~al.(2020)Kim, Yang, Kim, and Lee}]{kim2020somdst}
Sungdong Kim, Sohee Yang, Gyuwan Kim, and Sang-woo Lee. 2020.
\newblock Efficient dialogue state tracking by selectively overwriting memory.
\newblock In \emph{ACL}.

\bibitem[{Lee et~al.(2019{\natexlab{a}})Lee, Lee, and
  Kim}]{lee_etal_2019_sumbt}
Hwaran Lee, Jinsik Lee, and Tae-Yoon Kim. 2019{\natexlab{a}}.
\newblock \href {https://doi.org/10.18653/v1/P19-1546} {{SUMBT}: Slot-utterance
  matching for universal and scalable belief tracking}.
\newblock In \emph{Proceedings of the 57th Annual Meeting of the Association
  for Computational Linguistics}, pages 5478--5483, Florence, Italy.
  Association for Computational Linguistics.

\bibitem[{Lee et~al.(2019{\natexlab{b}})Lee, Zhu, Takanobu, Zhang, Zhang, Li,
  Li, Peng, Li, Huang, and Gao}]{lee_etal_2019_convlab}
Sungjin Lee, Qi~Zhu, Ryuichi Takanobu, Zheng Zhang, Yaoqin Zhang, Xiang Li,
  Jinchao Li, Baolin Peng, Xiujun Li, Minlie Huang, and Jianfeng Gao.
  2019{\natexlab{b}}.
\newblock \href {https://doi.org/10.18653/v1/P19-3011} {Convlab: Multi domain
  end-to-end dialog system platform}.
\newblock In \emph{Proceedings of the 57th Annual Meeting of the Association
  for Computational Linguistics: System Demonstrations}, pages 64--69,
  Florence, Italy. Association for Computational Linguistics.

\bibitem[{Mehri et~al.(2020)Mehri, Eric, and Hakkani-Tur}]{mehri2020dialoglue}
Shikib Mehri, Mihail Eric, and Dilek Hakkani-Tur. 2020.
\newblock \href {http://arxiv.org/abs/2009.13570} {Dialoglue: A natural
  language understanding benchmark for task-oriented dialogue}.

\bibitem[{Pan et~al.(2019)Pan, Bai, Wang, Zhou, and
  Liu}]{pan_etal_2019_improving}
Zhufeng Pan, Kun Bai, Yan Wang, Lianqiang Zhou, and Xiaojiang Liu. 2019.
\newblock \href {https://doi.org/10.18653/v1/D19-1191} {Improving open-domain
  dialogue systems via multi-turn incomplete utterance restoration}.
\newblock In \emph{Proceedings of the 2019 Conference on Empirical Methods in
  Natural Language Processing and the 9th International Joint Conference on
  Natural Language Processing (EMNLP-IJCNLP)}, pages 1824--1833, Hong Kong,
  China. Association for Computational Linguistics.

\bibitem[{Quan et~al.(2019)Quan, Xiong, Webber, and Hu}]{quan_etal_2019_gecor}
Jun Quan, Deyi Xiong, Bonnie Webber, and Changjian Hu. 2019.
\newblock \href {https://doi.org/10.18653/v1/D19-1462} {{GECOR}: An end-to-end
  generative ellipsis and co-reference resolution model for task-oriented
  dialogue}.
\newblock In \emph{Proceedings of the 2019 Conference on Empirical Methods in
  Natural Language Processing and the 9th International Joint Conference on
  Natural Language Processing (EMNLP-IJCNLP)}, pages 4547--4557, Hong Kong,
  China. Association for Computational Linguistics.

\bibitem[{Rastogi et~al.(2019)Rastogi, Zang, Sunkara, Gupta, and
  Khaitan}]{rastogi2019towards}
Abhinav Rastogi, Xiaoxue Zang, Srinivas Sunkara, Raghav Gupta, and Pranav
  Khaitan. 2019.
\newblock Towards scalable multi-domain conversational agents: The
  schema-guided dialogue dataset.
\newblock \emph{arXiv preprint arXiv:1909.05855}.

\bibitem[{Ren et~al.(2019)Ren, Ni, and McAuley}]{ren_etal_2019_scalable}
Liliang Ren, Jianmo Ni, and Julian McAuley. 2019.
\newblock \href {https://doi.org/10.18653/v1/D19-1196} {Scalable and accurate
  dialogue state tracking via hierarchical sequence generation}.
\newblock In \emph{Proceedings of the 2019 Conference on Empirical Methods in
  Natural Language Processing and the 9th International Joint Conference on
  Natural Language Processing (EMNLP-IJCNLP)}, pages 1876--1885, Hong Kong,
  China. Association for Computational Linguistics.

\bibitem[{Schatzmann et~al.(2007)Schatzmann, Thomson, Weilhammer, Ye, and
  Young}]{schatzmann2007agenda}
Jost Schatzmann, Blaise Thomson, Karl Weilhammer, Hui Ye, and Steve Young.
  2007.
\newblock Agenda-based user simulation for bootstrapping a pomdp dialogue
  system.
\newblock In \emph{Human Language Technologies 2007: The Conference of the
  North American Chapter of the Association for Computational Linguistics;
  Companion Volume, Short Papers}, pages 149--152.

\bibitem[{Su et~al.(2019)Su, Shen, Zhang, Sun, Hu, Niu, and
  Zhou}]{su_etal_2019_improving}
Hui Su, Xiaoyu Shen, Rongzhi Zhang, Fei Sun, Pengwei Hu, Cheng Niu, and Jie
  Zhou. 2019.
\newblock \href {https://doi.org/10.18653/v1/P19-1003} {Improving multi-turn
  dialogue modelling with utterance {R}e{W}riter}.
\newblock In \emph{Proceedings of the 57th Annual Meeting of the Association
  for Computational Linguistics}, pages 22--31, Florence, Italy. Association
  for Computational Linguistics.

\bibitem[{Takanobu et~al.(2019)Takanobu, Zhu, and
  Huang}]{takanobu_etal_2019_guided}
Ryuichi Takanobu, Hanlin Zhu, and Minlie Huang. 2019.
\newblock \href {https://doi.org/10.18653/v1/D19-1010} {Guided dialog policy
  learning: Reward estimation for multi-domain task-oriented dialog}.
\newblock In \emph{Proceedings of the 2019 Conference on Empirical Methods in
  Natural Language Processing and the 9th International Joint Conference on
  Natural Language Processing (EMNLP-IJCNLP)}, pages 100--110, Hong Kong,
  China. Association for Computational Linguistics.

\bibitem[{Wang et~al.(2020)Wang, Guo, and Zhu}]{wang_etal_2020_slot}
Yexiang Wang, Yi~Guo, and Siqi Zhu. 2020.
\newblock \href {https://www.aclweb.org/anthology/2020.emnlp-main.243} {Slot
  attention with value normalization for multi-domain dialogue state tracking}.
\newblock In \emph{Proceedings of the 2020 Conference on Empirical Methods in
  Natural Language Processing (EMNLP)}, pages 3019--3028, Online. Association
  for Computational Linguistics.

\bibitem[{Wen et~al.(2017)Wen, Vandyke, Mrk{\v{s}}i{\'c}, Ga{\v{s}}i{\'c},
  Rojas-Barahona, Su, Ultes, and Young}]{wen_etal_2017_network}
Tsung-Hsien Wen, David Vandyke, Nikola Mrk{\v{s}}i{\'c}, Milica
  Ga{\v{s}}i{\'c}, Lina~M. Rojas-Barahona, Pei-Hao Su, Stefan Ultes, and Steve
  Young. 2017.
\newblock A network-based end-to-end trainable task-oriented dialogue system.
\newblock In \emph{Proceedings of the 15th Conference of the {E}uropean Chapter
  of the Association for Computational Linguistics: Volume 1, Long Papers}.

\bibitem[{Williams et~al.(2013)Williams, Raux, Ramachandran, and
  Black}]{williams_etal_2013_dialog}
Jason Williams, Antoine Raux, Deepak Ramachandran, and Alan Black. 2013.
\newblock \href {https://www.aclweb.org/anthology/W13-4065} {The dialog state
  tracking challenge}.
\newblock In \emph{Proceedings of the {SIGDIAL} 2013 Conference}, pages
  404--413, Metz, France. Association for Computational Linguistics.

\bibitem[{Wu et~al.(2019)Wu, Madotto, Hosseini-Asl, Xiong, Socher, and
  Fung}]{wu_etal_2019_transferable}
Chien-Sheng Wu, Andrea Madotto, Ehsan Hosseini-Asl, Caiming Xiong, Richard
  Socher, and Pascale Fung. 2019.
\newblock \href {https://doi.org/10.18653/v1/P19-1078} {Transferable
  multi-domain state generator for task-oriented dialogue systems}.
\newblock In \emph{Proceedings of the 57th Annual Meeting of the Association
  for Computational Linguistics}, pages 808--819, Florence, Italy. Association
  for Computational Linguistics.

\bibitem[{Zang et~al.(2020)Zang, Rastogi, and Chen}]{zang_etal_2020_multiwoz}
Xiaoxue Zang, Abhinav Rastogi, and Jindong Chen. 2020.
\newblock \href {https://www.aclweb.org/anthology/2020.nlp4convai-1.13}
  {{M}ulti{WOZ} 2.2 : A dialogue dataset with additional annotation corrections
  and state tracking baselines}.
\newblock In \emph{Proceedings of the 2nd Workshop on Natural Language
  Processing for Conversational AI}, pages 109--117, Online. Association for
  Computational Linguistics.

\bibitem[{Zhang et~al.(2019)Zhang, Hashimoto, Wu, Wan, Yu, Socher, and
  Xiong}]{zhang2019find}
Jian-Guo Zhang, Kazuma Hashimoto, Chien-Sheng Wu, Yao Wan, Philip~S Yu, Richard
  Socher, and Caiming Xiong. 2019.
\newblock Find or classify? dual strategy for slot-value predictions on
  multi-domain dialog state tracking.
\newblock \emph{arXiv preprint arXiv:1910.03544}.

\bibitem[{Zhao et~al.(2019)Zhao, Xie, and Eskenazi}]{zhao_etal_2019_rethinking}
Tiancheng Zhao, Kaige Xie, and Maxine Eskenazi. 2019.
\newblock \href {https://doi.org/10.18653/v1/N19-1123} {Rethinking action
  spaces for reinforcement learning in end-to-end dialog agents with latent
  variable models}.
\newblock In \emph{Proceedings of the 2019 Conference of the North {A}merican
  Chapter of the Association for Computational Linguistics: Human Language
  Technologies, Volume 1 (Long and Short Papers)}, pages 1208--1218,
  Minneapolis, Minnesota. Association for Computational Linguistics.

\bibitem[{Zhou and Small(2019)}]{zhou2019multi}
Li~Zhou and Kevin Small. 2019.
\newblock Multi-domain dialogue state tracking as dynamic knowledge graph
  enhanced question answering.
\newblock \emph{arXiv preprint arXiv:1911.06192}.

\bibitem[{Zhu et~al.(2020{\natexlab{a}})Zhu, Huang, Zhang, Zhu, and
  Huang}]{zhu2020crosswoz}
Qi~Zhu, Kaili Huang, Zheng Zhang, Xiaoyan Zhu, and Minlie Huang.
  2020{\natexlab{a}}.
\newblock Crosswoz: A large-scale chinese cross-domain task-oriented dialogue
  dataset.
\newblock \emph{arXiv preprint arXiv:2002.11893}.

\bibitem[{Zhu et~al.(2020{\natexlab{b}})Zhu, Zhang, Fang, Li, Takanobu, Li,
  Peng, Gao, Zhu, and Huang}]{zhu_etal_2020_convlab}
Qi~Zhu, Zheng Zhang, Yan Fang, Xiang Li, Ryuichi Takanobu, Jinchao Li, Baolin
  Peng, Jianfeng Gao, Xiaoyan Zhu, and Minlie Huang. 2020{\natexlab{b}}.
\newblock \href {https://doi.org/10.18653/v1/2020.acl-demos.19} {{C}onv{L}ab-2:
  An open-source toolkit for building, evaluating, and diagnosing dialogue
  systems}.
\newblock In \emph{Proceedings of the 58th Annual Meeting of the Association
  for Computational Linguistics: System Demonstrations}, pages 142--149,
  Online. Association for Computational Linguistics.

\end{thebibliography}
\bibliographystyle{acl_natbib}

\clearpage
\appendix
\onecolumn
\section{Value Normalization}
\label{sec:appendix}
\begin{table}[!htb]

\begin{center}
    \begin{tabularx}{\textwidth}{|c|c|}
    \hline
    \textbf{Type} & \textbf{Content}\\
    \hline
    Number & \begin{tabular}{X}
    {'zero': '0', 'one': '1', 'two': '2', 'three': '3', 'four': '4', 'five': '5', 'six': '6', 'seven': '7', 'eight': '8', 'nine': '9', 'ten': '10', 'eleven': '11', 'twelve': '12'}\\      
    \end{tabular} \\
    \hline
    
    Pricerange & \begin{tabular}{X}
    {'high end': 'expensive', 'expensively': 'expensive', 'upscale': 'expensive', 'inexpensive': 'cheap', 'cheaply': 'cheap', 'cheaper': 'cheap', 'cheapest': 'cheap', 'moderately priced': 'moderate', 'moderately': 'moderate'}
    \end{tabular} \\
    \hline
    
    dontcare & \begin{tabular}{X}
    {'do n't have a preference': 'dontcare', 'do not have a preference': 'dontcare', 'no particular': 'dontcare', 'not particular': 'dontcare', 'do not care': 'dontcare', 'do n't care': 'dontcare', 'any': 'dontcare', 'does not matter': 'dontcare', 'does n't matter': 'dontcare', 'not really': 'dontcare', 'do nt care': 'dontcare', 'does n really matter': 'dontcare', 'do n't really care': 'dontcare'}
    \end{tabular} \\
    \hline
    
    Area & \begin{tabular}{X}
    'center': 'centre', 'northern': 'north', 'northside': 'north', 'eastern': 'east' 'eastside': 'east', 'westside': 'west', 'western': 'west', 'southside': 'south', 'southern': 'south' 
    \end{tabular} \\
    \hline
    
    Time & \begin{tabular}{X}
    Remove words as 'after', 'before' and etc., and sort to the 'hh:mm' time format. 'X pm' format is remained as the original.
    \end{tabular} \\
    \hline
    
    Stars & \begin{tabular}{X}
    [0-9]-stars, converted to [0-9] 
    \end{tabular} \\
    \hline
    
    Parking and Internet & \begin{tabular}{X}
    'Free' value for parking and internet slot is converted to 'yes'
    \end{tabular} \\
    \hline
    
    Plural & \begin{tabular}{X}
    'hotels': 'hotel', 'guesthouses': 'guesthouse', 'churches': 'church', 'museums': 'museum', 'entertainments': 'entertainment', 'colleges': 'college', 'nightclubs': 'nightclub', 'swimming pools': 'swimming pool', 'architectures': 'architecture', 'cinemas': 'cinema', 'boats': 'boat', 'boating': 'boat', 'theatres': 'theatre', 'concert halls': 'concert hall', 'parks': 'park', 'local sites': 'local site', 'hotspots': 'hotspot'
    \end{tabular} \\
    \hline
    
    \end{tabularx}

    \caption{Value normalization rules when updating values from dialogue acts to dialogue states.}
    \label{DAtoState}
\end{center}
\end{table}

\section{Dialogue State Tracking benchmarks}
\label{sec:appendix2}
Upon code availability, we experiment MultiWOZ 2.3 on various dialogue state tracking models and Table \ref{MultipleDST} shows the corresponding joint goal accuracies.
\begin{table}[h!]
\centering
\begin{tabular}{|c|c|c|}
\hline
    \textbf{Models} & \textbf{MultiWOZ 2.1} & \textbf{MultiWOZ 2.3} \\
    \hline
    TRADE~\citep{wu_etal_2019_transferable} & 45.6\% & \textbf{49.2}\%  \\
    \hline
    SUMBT~\citep{lee_etal_2019_sumbt} & 49.2\% & \textbf{52.9}\% \\
    \hline
    COMER~\citep{ren_etal_2019_scalable} & 48.8\% & \textbf{50.2}\% \\
    \hline
    DSTQA~\citep{zhou2019multi} & 51.2\% & \textbf{51.8}\% \\
    \hline
    SOM-DST~\citep{kim2020somdst} & 53.1\% & \textbf{55.5}\% \\
    \hline
    TripPy~\citep{heck_etal_2020_trippy} & 55.3\% & \textbf{63.0}\% \\
    \hline
    ConvBERT-DG-Multi~\citep{mehri2020dialoglue} & 58.7\% &  \textbf{67.9}\% \\
    \hline
    SAVN~\citep{wang_etal_2020_slot} & 54.5\% & \textbf{58.0}\% \\
    \hline
\end{tabular}
\caption{Joint goal accuracies for different dialogue state tracking models on the MultiWOZ 2.1 and MultiWOZ-coref. We notice our work is cocurrent with MultiWOZ 2.2. However, we mainly base our refinement on MultiWOZ 2.1 and many models do not report joint goal accuracies on MultiWOZ 2.2. Therefore, MultiWOZ 2.2 is excluded from comparison. }
\label{MultipleDST}
\end{table}

\section{SUMBT Slot Accuracy}
\label{sec:appendix1}

\begin{table*}[!htb]
\centering
\begin{tabular}{|l|l|l|l|l|}
\hline
\textbf{Slot type}  & \textbf{MultiWOZ 2.1}   & \textbf{MultiWOZ 2.2}   & \textbf{MultiWOZ 2.3}   & \textbf{MultiWOZ-coref} \\ \hline
attraction-area        & 95.94          & 95.97          & \textbf{96.28} & \color{red}\textbf{96.80} \\ \hline
attraction-name        & 93.64          & 93.92          & \color{red}\textbf{95.28} & \textbf{94.59} \\ \hline
attraction-type        & 96.76          & \textbf{97.12} & 96.53          & 96.91          \\ \hline
hotel-area             & 94.33          & 94.44          & \textbf{94.65} & \color{red}\textbf{95.02} \\ \hline
hotel-book day         & 98.87          & 99.06          & 99.04          & \color{red}\textbf{99.32} \\ \hline
hotel-book people      & 98.66          & 98.72          & \textbf{98.93} & \color{red}\textbf{99.17} \\ \hline
hotel-book stay        & 99.23          & 99.50          & \color{red}\textbf{99.70} & \color{red}\textbf{99.70} \\ \hline
hotel-internet         & 97.02          & 97.02          & \textbf{97.45} & \color{red}\textbf{97.56} \\ \hline
hotel-name             & 94.67          & 93.76          & \color{red}\textbf{94.71} & \color{red}\textbf{94.71} \\ \hline
hotel-parking          & 97.04          & 97.19          & \textbf{97.90} & \color{red}\textbf{98.34} \\ \hline
hotel-pricerange       & 96.00          & 96.23          & 95.90          & \color{red}\textbf{96.40} \\ \hline
hotel-stars            & 97.88          & 97.95          & \textbf{97.99} & \color{red}\textbf{98.09} \\ \hline
hotel-type             & 94.67          & 94.22          & \color{red}\textbf{95.92} & \textbf{95.65} \\ \hline
restaurant-area        & \textbf{96.30} & 95.47          & 95.52          & 96.05          \\ \hline
restaurant-book day    & 98.90          & 98.91          & 98.83          & \color{red}\textbf{99.66} \\ \hline
restaurant-book people & 98.91          & 98.98          & \textbf{99.17} & \color{red}\textbf{99.21} \\ \hline
restaurant-book time   & 99.43          & 99.24          & 99.31          & \color{red}\textbf{99.46} \\ \hline
restaurant-food        & \textbf{97.69} & 97.61          & 97.49          & 97.64          \\ \hline
restaurant-name        & 92.71          & 93.18          & \color{red}\textbf{95.10} & \textbf{94.91} \\ \hline
restaurant-pricerange  & 95.36          & 95.65          & \textbf{95.75} & \color{red}\textbf{96.26} \\ \hline
taxi-arriveBy          & 98.36          & 98.03          & 98.18          & \color{red}\textbf{98.45} \\ \hline
taxi-departure         & 96.13          & 96.35          & 96.15          & \color{red}\textbf{97.49} \\ \hline
taxi-destination       & 95.70          & 95.50          & 95.56          & \color{red}\textbf{97.59} \\ \hline
taxi-leaveAt           & 98.91          & 98.96          & \color{red}\textbf{99.04} & \textbf{99.02} \\ \hline
train-arriveBy         & 96.40          & 96.40          & \textbf{96.54} & \color{red}\textbf{96.76} \\ \hline
train-book people      & 97.26          & 97.04          & \textbf{97.29} & \color{red}\textbf{97.67} \\ \hline
train-day              & 98.63          & 98.60          & \textbf{99.04} & \color{red}\textbf{99.38} \\ \hline
train-departure        & \textbf{98.43} & 98.40          & 97.56          & 97.50          \\ \hline
train-destination      & \textbf{98.55} & 98.30          & 97.96          & 97.86          \\ \hline
train-leaveAt          & 93.64          & \textbf{94.14} & 93.98          & 93.96          \\ \hline
\end{tabular}
\caption{Slot accuracies among MultiWOZ 2.1, MultiWOZ 2.2, MultiWOZ 2.3 and MultiWOZ-coref in terms of different slot types. The bold number indicates the highest accuracy across all three datasets for each slot. The red bold number indicates higher accuracy between MultiWOZ 2.3 and MultiWOZ-coref for each slot.}
\label{tab:sumbt-merge}
\end{table*}













\end{document}